%% file: root.tex
\documentclass[letterpaper, 10 pt, conference]{ieeeconf}  

\IEEEoverridecommandlockouts                              

\usepackage{color}
\usepackage{xcolor}
\usepackage{graphicx}
\usepackage{amsmath}
\usepackage{hyperref}
\usepackage{xurl}

\title{\LARGE \bf
LRDDv3: High-Resolution Long-Range Drone Detection Dataset with Range Information and Thermal Data
}

\author{Knut Peterson$^{1*}$, Zaid Mayers$^{1}$, Azmain Yousuf$^{1}$, Priontu Chowdhury$^{1}$, Asher Zaczepinski$^{1}$,
\\ Solmaz Arezoomandan$^{1}$, 
Reihaneh Maarefdoust$^{2}$, and David Han$^{1}$ 
\thanks{*Corresponding author {\tt\small knut.peterson@drexel.edu} }
\thanks{$^{1}$iMaPLe Research Lab, Drexel University}%
\thanks{$^{2}$University of Maine}%
}

\begin{document}

\maketitle
\thispagestyle{empty}
\pagestyle{empty}

\begin{abstract}

Unmanned Aerial Vehicles (UAVs) have quickly become common in various airspaces, representing a wide range of applications from recreation flying to commercial photography and package delivery. With the increasing prevalence of UAVs, it becomes critical that both manned and unmanned aircraft can detect UAVs and other flying objects from long range to effectively track movement and ensure safe operation in shared spaces. While several datasets have been introduced for drone detection, the need for expanded high-quality data persists, especially in the area of high-resolution long-range drone data. To address this, we introduce a high-resolution dataset of 102,532 long-range RGB images of drones, sampled at 5 FPS from 128 distinct video clips taken mid flight during 17 different data collection days spread over 8 months to ensure a wide variety of lighting scenarios, flight locations, and background elements. The dataset boasts comprehensive drone range information across the dataset, as well as 29,630 IR images, all paired with RGB counterparts from the base dataset. As one of the first drone detection datasets to leverage 4K image resolution and paired 640x512 IR images, our work represents a significant advancement to enable the detection of drones at long range.
For access to the complete dataset, please visit 
\url{https://research.coe.drexel.edu/ece/imaple/lrddv3/}.

\end{abstract}


\input{sections/1.Introduction}
\input{sections/2.Related}

\input{sections/3.Dataset}

\input{sections/4.Benchmark}

\input{sections/5.Conclusion}

\addtolength{\textheight}{-4cm}






\bibliographystyle{ieeetr}
\bibliography{ref.bib}

\end{document}

%% file: sections/1.Introduction.tex
\section{Introduction}


The use of Unmanned Aerial Vehicles (UAVs) has been exponentially growing for a wide range of applications from recreation flying to commercial photography and package delivery in both urban and rural environments \cite{mohsan2023unmanned, karachalios_2022_euromicro}. 
With the use of UAVs becoming more prevalent, it becomes critical that aerial systems can detect UAVs and other flying objects from long range to effectively track movement and avoid collisions. Different manned or unmanned aircraft need to be able to detect UAVs as we are heading toward a future where both must operate together, sometimes in shared spaces. Various advanced detection systems rely on deep learning algorithms, which can be significantly affected by the size and quality of the datasets used in their training. However, many existing datasets showcase drone images that do not reflect the context in which they will be encountered when being operated for commercial or recreational purposes, like varying weather conditions, backgrounds, time of the day, obstacles seen in dense urban environments \cite{rouhi2024_LRDD, rouhi2024_LRDDv2}.

\begin{figure*}[!tb] 
\centering
\includegraphics[width=0.9\textwidth]{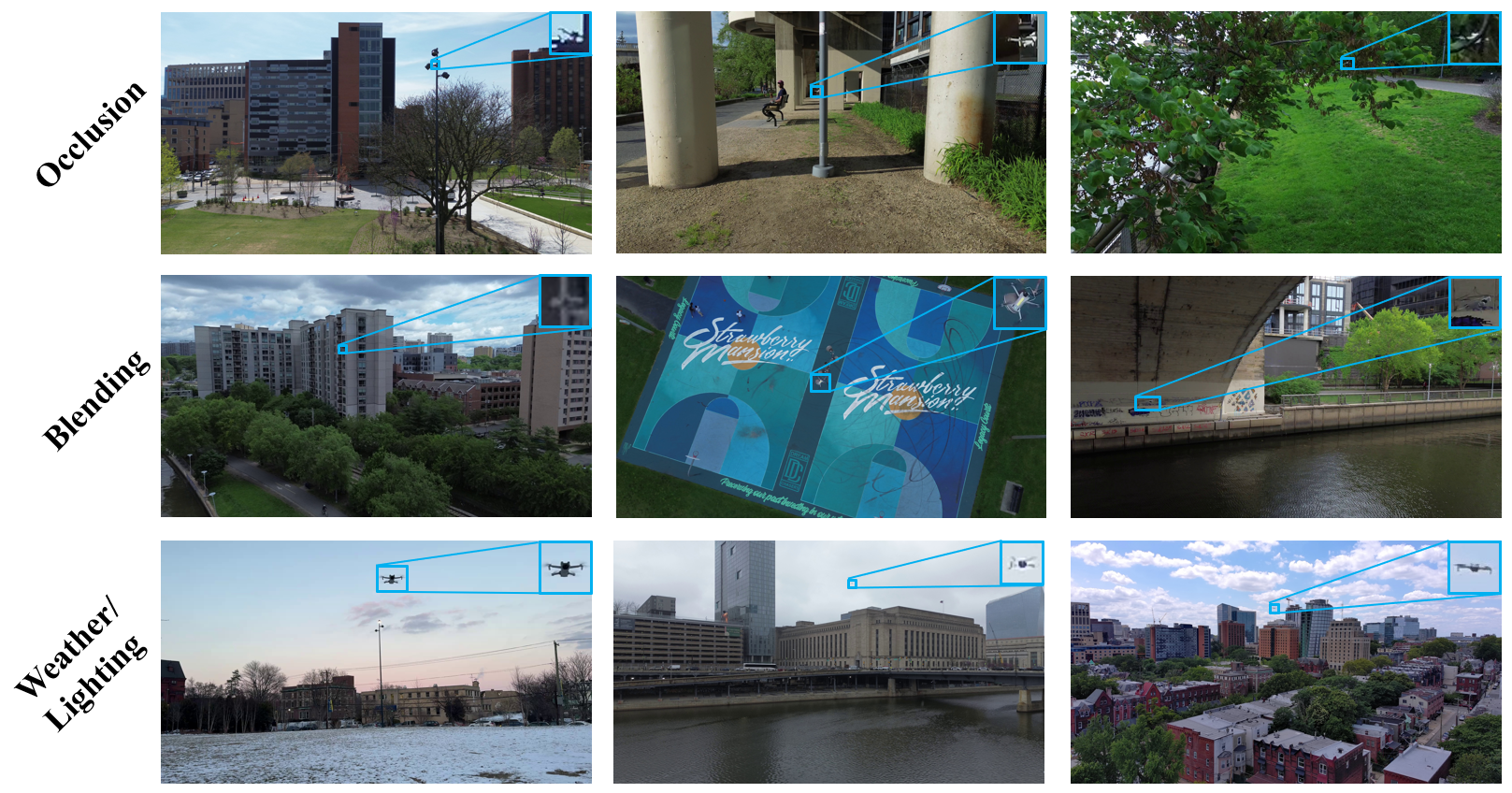}
\caption{Examples of images in the dataset, featuring different challenges such as drones occluded by objects and foliage, drones blending into noisy backgrounds, and drones flying in snowy, rainy, and clear weather conditions.}
\label{fig:challenges}
\end{figure*}

To bridge this gap, we introduce LRDDv3: High-Resolution Long-Range Drone Detection Dataset. Consisting of 102,532 RGB images taken mid flight during 17 different data collection days spread over 8 months, 29,630 IR images paired with RGB counterparts from the base dataset, and comprehensive drone range information, LRDDv3 stands out for its scale, high-resolution, and image diversity. Additionally, the dataset includes a variety of challenges and features to aid in model generalization and enable robust detections, detailed as follows:

\textbf{Thermal Imaging:} UAVs have a heat signature that can vary from the elements of their background, such as buildings, trees, cars, sky, etc \cite{fudala_2019}. Using thermal imaging, either on its own or in combination with RGB imaging, drone detection can be made more robust through differentiating between the heat signatures of the drone and background elements.

\textbf{Illumination Variation:} Change of lighting conditions can significantly change the appearance of drones seen from another drone during a flight \cite{Munir_2024_WACV, zhong_2016_spie}. This dataset incorporates images taken in varying backgrounds at different times of the day. i.e. dawn, dusk, noon etc. to reflect real world scenarios in drone operations at different times of the day. The variation in illumination level and shadow effects aim to better train algorithms to recognize drones at different times of the day.

\textbf{Diverse Backgrounds:} UAVs are expected to operate in a wide range of environments from dense cities to sparsely inhabited small towns or recreational areas. Advanced drone detection systems need to pick up on the varying contexts of the backgrounds and adapt to different background textures, components and complexities \cite{Munir_2024_WACV, rouhi2024_LRDD, rouhi2024_LRDDv2}. By incorporating images where drones are operating in a dense urban areas or in a more open parks or over a water body, LRDDv3 aims to aid in improving detection accuracy across varying contexts.

\textbf{Weather Conditions:} Changes in weather conditions affect UAV visibility and add new contexts for detection \cite{singh_2024_icdt, rouhi2024_LRDD, rouhi2024_LRDDv2}. Our work includes images with a wide range of weather conditions from clear skies to grey-cloudy conditions and urban snowy landscapes.

\textbf{Occlusion and Background Blending:} While flying in dense urban areas, UAVs often blend into the background or become partially or fully occluded by objects, such as buildings, traffic lights and road signs, cars and other vehicles etc. \cite{wang_2022_mdpi, rouhi2024_LRDDv2} We address this by including images showing scenarios where the drone is partially indistinguishable from its immediate background, testing any algorithm’s capability in drone detection under compromised visibility.

\textbf{Long-Range:}
As UAV technology has progressed, the speed of UAVs has also increased, making it essential for detection systems to be capable of detecting them at long range \cite{khorshand_2025_icce}. To aid with this, the dataset includes images of drones at distances from 0-200 meters from the camera, along with range metadata. The resulting instances of drones in the dataset include bounding boxes with areas as small as 12 total pixels, posing a difficult challenge in long-range drone detection.

LRDDv3 improves on previous works by combining 4K resolution, IR imaging, drone range annotations, mobile camera views, and diverse lighting, weather, and background conditions into a single, comprehensive dataset. Our work seeks to significantly improve the data quality of drone datasets with diverse images taken in varying scenarios that aim to mimic conditions in which drone detection will be necessary for in-flight operations in the real world, enabling more reliable integration of drones in national and international airspaces.

%% file: sections/2.Related.tex
\section{Related Work - Existing Datasets}

\begin{table*}
\caption{Overview of current drone detection datasets}
\label{table:datasets}
\resizebox{\textwidth}{!}{%
\renewcommand{\arraystretch}{1.15}
\begin{tabular}{|l|l|l|l|l|l|l|l|l|}
\hline
\textbf{Dataset} &
  \textbf{\# of Images/Videos} &
  \textbf{Data Type} &
  \textbf{Resolution} &
  \textbf{Camera Dynamics} &
  \textbf{Lighting} &
  \textbf{Occlusion} &
  \textbf{Weather} &
  \textbf{Range} \\ \hline
Multi-Sensor &
  \begin{tabular}[c]{@{}l@{}}203k images\\ 90 audio clips\end{tabular} &
  \begin{tabular}[c]{@{}l@{}}RGB,\\ IR,\\ Audio\end{tabular} &
  \begin{tabular}[c]{@{}l@{}}640x512 (RGB)\\ 320x256 (IR)\end{tabular} &
  Static &
  No &
  No &
  Partial &
  Categorical \\ \hline
BirDrone &
  3.3k images &
  RGB &
  640x640 &
  \begin{tabular}[c]{@{}l@{}}Various (from\\ internet)\end{tabular} &
  Yes &
  No &
  No &
  No \\ \hline
VisoDECT &
  20.9k images &
  \begin{tabular}[c]{@{}l@{}}RGB,\\ EO-IR\end{tabular} &
  852x480 &
  Static/Handheld &
  Yes &
  No &
  Yes &
  No \\ \hline
RWOQ &
  56.8k images &
  RGB &
  640x480 &
  \begin{tabular}[c]{@{}l@{}}Various (from\\ internet)\end{tabular} &
  Yes &
  Partial &
  Yes &
  No \\ \hline
SynDroneVision &
  131.3k images &
  RGB &
  2560x1489 &
  Mobile/Flying &
  Yes &
  No &
  No &
  No \\ \hline
DUT Anti-UAV &
  10k images &
  RGB &
  \begin{tabular}[c]{@{}l@{}}Variable (240x160\\ to 5616x3744)\end{tabular} &
  Various &
  Yes &
  No &
  Partial &
  No \\ \hline
Drone Hunter &
  58.6k images &
  RGB &
  1280x720 &
  Handheld &
  Partial &
  No &
  No &
  No \\ \hline
FRED &
  \begin{tabular}[c]{@{}l@{}}700k images (30\\FPS from video)\end{tabular} &
  \begin{tabular}[c]{@{}l@{}}RGB,\\ Event\end{tabular} &
  1280x720 &
  Static &
  Yes &
  No &
  Yes &
  No \\ \hline
ARD-MAV &
  \begin{tabular}[c]{@{}l@{}}107.5k images (30\\FPS from 60 videos)\end{tabular} &
  RGB &
  1920x1080 &
  Mobile/Flying &
  No &
  Yes &
  No &
  No \\ \hline
\begin{tabular}[c]{@{}l@{}}Multi-View\\ Drone Tracking\end{tabular} &
  5 videos, 30.5min &
  RGB &
  \begin{tabular}[c]{@{}l@{}}Variable\\ (1440x1080 to 4K)\end{tabular} &
  \begin{tabular}[c]{@{}l@{}}Static\\ (Multi-camera)\end{tabular} &
  No &
  No &
  Partial &
  No \\ \hline
UAV-Detect &
  20.4k images &
  RGB &
  640x640 &
  Mobile/Flying &
  Yes &
  No &
  No &
  No \\ \hline
Det-Fly &
  13.3k images &
  RGB &
  4K &
  Mobile/Flying &
  Yes &
  Yes &
  No &
  No \\ \hline
Drone vs Bird &
\begin{tabular}[c]{@{}l@{}}97.4k images (30FPS\\from 77 videos)\end{tabular}&
  RGB &
  \begin{tabular}[c]{@{}l@{}}Variable\\ (4K, 1920x1080)\end{tabular} &
  Various &
  No &
  Yes &
  No &
  No \\ \hline
LRDDv1 &
  \begin{tabular}[c]{@{}l@{}}21.2k images\\ (10FPS sampling)\end{tabular} &
  RGB &
  1920x1080 &
  Mobile/Flying &
  Yes &
  Yes &
  Yes &
  No \\ \hline
LRDDv2 &
  \begin{tabular}[c]{@{}l@{}}17k images\\ (10FPS sampling)\end{tabular} &
  RGB &
  1920x1080 &
  Mobile/Flying &
  Yes &
  Yes &
  Yes &
  Partial \\ \hline
LRDDv3 &
  \begin{tabular}[c]{@{}l@{}}102k RGB, 29.6k IR\\ (5FPS sampling\\ from 128 videos)\end{tabular} &
  \begin{tabular}[c]{@{}l@{}}RGB,\\ IR\end{tabular} &
  \begin{tabular}[c]{@{}l@{}}4K (RGB)\\ 640x512 (IR)\end{tabular} &
  Mobile/Flying &
  Yes &
  Yes &
  Yes &
  Yes \\ \hline
\end{tabular}
}
\end{table*}


To date, a variety of different datasets have been introduced to approach the problem of drone detection. Many of these datasets are tailored for specific purposes and are examined as follows, with an overview of different datasets and their features shown in Table \ref{table:datasets}.

The \textbf{Multi-Sensor Drone Detection dataset} \cite{svanstrom_2021_multisensor} tackles the critical need for multi-modal drone detection by using infrared/visible videos along with audio files. The video data contains 365 infrared (320×256 pixels) and 285 visible (640×512 pixels) 10 second+ videos of drones, birds, airplanes, and helicopters resulting in 203,328 annotated frames. Additionally, it includes 90 audio files recording drones, helicopters, and background noise. The dataset is split into three categories (Close, Medium, Distant), with drone detection ranges reaching up to 200m.

The \textbf{BirDrone dataset} \cite{zamri_2024_birdrone} specifically targets the challenge of drone vs. bird classification with 2,970 640x640 resolution images, featuring 2,617 drone images and 353 bird images. These images have complex background and lighting conditions to prepare machine learning models for real-world scenarios. The dataset includes 6,162 annotations with bounding boxes ranging from 7x14 to 65x182 pixels.

\textbf{VisioDECT} \cite{islam2025visiodect} is a drone dataset used for countering unauthorized drone deployments using visual and electro-optical infrared detection technologies. The dataset contains 20,924 images (852x480 pixel resolution) and annotations from 6 drone models spanning 3 scenarios (cloudy, sunny, and evening), at varying altitudes and distances (30m-100m). The dataset was collected at 12 different locations over a period of 1 year and 8 months.

The \textbf{Real World Object Detection Dataset for Quadcopter UAV
Detection} (RWOQ) \cite{rwoq_2020} includes drones in many types, sizes, positions, and environments. This variety of drones focuses specifically on quadcopter detection. The dataset consists of 51,336 training images and 5,375 test RGB images at 640x480 resolution.

The \textbf{SynDroneVision dataset} \cite{lenhard_2025_syndronevision} presents an innovative approach to drone dataset creation through synthetic data generation. This data features diverse environments, drone models, and lighting conditions specifically designed for RGB-based drone detection in surveillance applications. This unique dataset addresses the scarcity of large-scale annotated training data while significantly reducing acquisition costs. However, the sim-to-real transfer has yet to be fully addressed, so there is still  work to be done before we can fully rely on synthetic data for drone detection.

The \textbf{DUT Anti-UAV dataset} \cite{zhao_2022_antiuav} is one of the best single-modal drone detection datasets with 10,000 drone images separated into 5,200 for training, 2,600 for validation, and 2,200 for testing. It uses 35 different drone models across various backgrounds, lighting conditions, and weather scenarios. The dataset is distinguished by its high diversity and has the drones captured in (day, night, dawn, dusk) and on (sunny, cloudy, snowy) days. It is composed of a wide variety of image resolutions, ranging from 240x160 to 5616x3744 pixels. On average, drones in the dataset occupy less than 5\% of the image area. Also, the dataset has a tracking component with 20 videos that have short and long term sequences.

The \textbf{Drone Hunter dataset} \cite{wyder_2019_dronehunter} was developed for autonomous UAVs that can detect and neutralize other drones. The dataset comprises 58,647 images, 10,000 of which are synthetically generated from AirSim Simulator (similar to SynDroneVision). The real images are captured using drone-to-drone footage captured using specialized equipment. This unique dataset helps build on the critical gap in counter-drone applications.

The \textbf{Florence RGB-Event Drone Dataset} (FRED) \cite{magrini_2025_fred} contains 700K+ frames spanning over 7 hours and features 5 different drone models captured under various weather challenges. The dataset combines traditional RGB sensing with event-based cameras to provide comprehensive temporal information. This unique approach enables motion blur-free tracking of fast-moving drones and allows precise trajectory prediction even in challenging lighting conditions where traditional cameras struggle. However, frames are not subsampled from video, and the cameras remain static during collection, so the dataset suffers significantly from limited background variation.

\begin{figure*}[!tb] 
\centering
\includegraphics[width=0.9\textwidth]{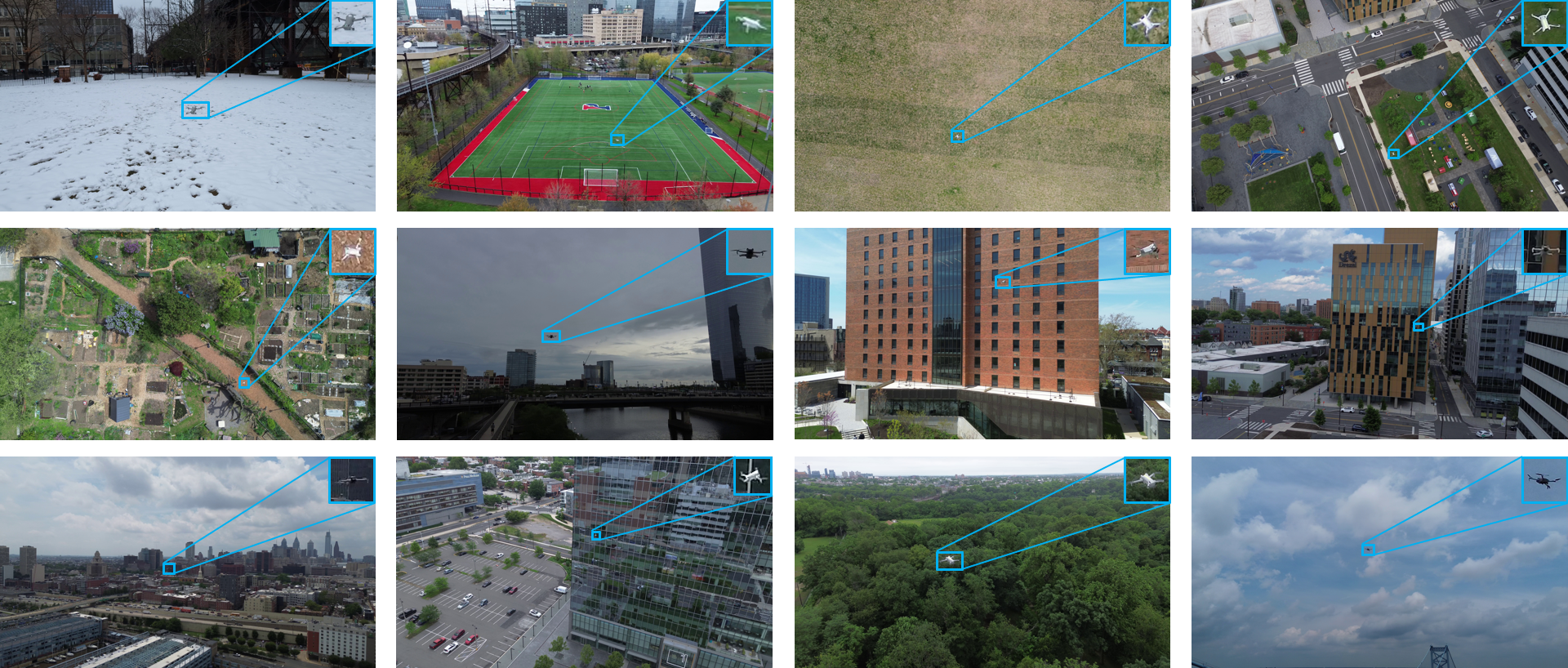}
\caption{Example images from the dataset, featuring a wide variety of backgrounds, camera angles, drone distances, and lighting conditions.}
\label{fig:backgrounds}
\end{figure*}

The \textbf{ARD-MAV dataset} \cite{guo2023globallocal}, introduced by Guo et al. in their work on global-local MAV detection, is one of the most prominent resources for evaluating air-to-air drone detection. It consists of approximately 107,500 high-resolution RGB images (1920×1080), sampled at 30 FPS from 60 video sequences captured using mobile, flying cameras. By leveraging onboard viewpoints, ARD-MAV introduces realistic challenges such as motion blur, scale variation, and cluttered backgrounds that static, ground-based datasets do not capture. Although it does not incorporate multimodal sensing (e.g., thermal imagery) or range annotations, ARD-MAV remains an important benchmark for testing algorithms under dynamic aerial conditions, serving as a bridge between controlled laboratory datasets and real-world drone-to-drone vision scenarios.

The \textbf{Multi-View Drone Tracking dataset} \cite{li_2020_multiview_drone_tracking} captures drone movements in 3D using several cameras positioned at different viewpoints. This multi perspective approach enables advanced tracking algorithms. However, it is limited by its use of consumer-grade cameras and because it only focuses on specific drone types.

The \textbf{UAV Detect dataset} \cite{uav-detect-pfiqs_dataset} consists of 20,400 images at 640x640 resolution with a wide variety of different backgrounds in both urban and rural areas, and different lighting conditions including challenges such as dim lighting and glare.

The \textbf{Det-Fly dataset} \cite{zheng_2021_detfly} focuses on air-to-air UAV detection scenarios with extremely small targets ranging from 1x1 to 149x95 pixels. The dataset consists of 13.3k images, captured at 4K resolution (3840×2160 pixels). Recent evaluations using enhanced models such as EA-DINO had Det-Fly achieving mAP 50 scores of 96.6\%. This demonstrates the datasets utility for evaluating small targets in aerial contexts.

The \textbf{Drone vs. Bird Detection Grand Challenge dataset} \cite{coluccia_2024_drone_vs_bird} has served as a benchmark for evaluating detection algorithms' potential to distinguish between drones and birds. This distinction serves a massive capability for reducing false positives in applications. The Drone vs. Bird dataset has been updated many times with the most recent version displayed in the 8th WOSDETC challenge at IJCNN 2025. This version focuses on small drone detection in complex environments and includes 77 videos with over 95k frames.

\begin{figure*}[!tb] 
\centering
\includegraphics[width=0.98\textwidth]{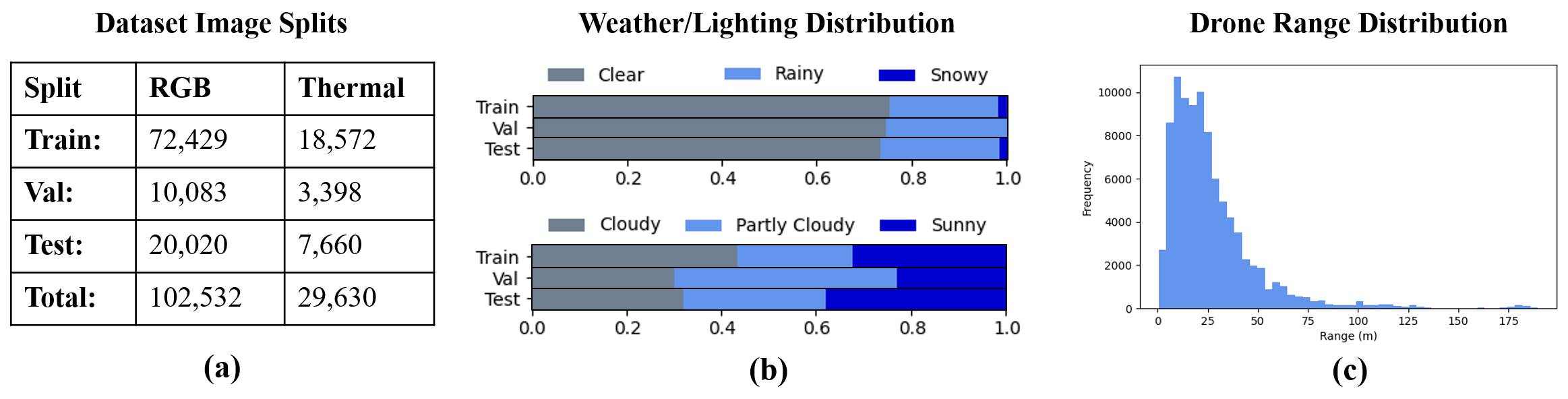}
\caption{An overview of dataset splits and distributions. (a) Shows the number of images in the training, validation, and test sets for both RGB and Thermal data. (b) Shows the weather (top) and lighting (bottom) distributions across the dataset. The dataset was primarily collected in clear weather, with some rainy and snowy conditions, while the lighting conditions were more balanced between cloudy, partly cloudy, and sunny. (c) Shows the distribution of the range between the target drone and the camera drone across the dataset. Most images were taken with 0-50 meters between the drones, but the dataset also includes longer ranges of up to 200 meters.}
\label{fig:distributions}
\end{figure*}

The \textbf{Long-Range Drone Detection Dataset} (LRDDv1 – 2024) \cite{rouhi2024_LRDD} was developed to bridge the quality gap in long-range drone detection by having training images captured in a variety of weather conditions, lighting scenarios, and different backgrounds at extended distances.. These images also included a wide range of drone positions and flight patterns. LRDDv2 (2025) \cite{rouhi2024_LRDDv2} expanded the original dataset to a total 39,516 annotated images (1920x1080 pixel resolution) and the standout feature being drone range data for over 8,000 images— a capability not present in any other dataset at the time of its release.

Datasets such as the \textbf{Airborne Object Tracking dataset} (AOT) \cite{AOT}, the \textbf{AMD3IR} dataset \cite{aqsa_2025_amd3ir}, and \textbf{EV-Flying} dataset \cite{Magrini_2025_ev_flying} are also relevant in the field of drone detection, but do not include RGB images, instead focusing on 8-bit grayscale (AOT), short-wave and long-wave IR (AMD3IR), and Event-based data (EV-Flying). Other datasets, such as the \textbf{U2U-D\&TD} dataset \cite{li_2022_u2udtd} and \textbf{MMFW-UAV} dataset \cite{MMFW-UAV} only focus on fixed-wing UAVs which represents a different challenge from our work. Additionally, other works such as the \textbf{DAC-SDC (Design Automation Conference System Design Content)} dataset \cite{xu_2021_dacsdc} and the  \textbf{UG2 Dataset} \cite{vidal2018ug} include annotated drones in their data, but they only consist of a relatively small portion of the overall data.

While many of these datasets exemplify specific areas of the drone detection landscape, none are positioned to include large dataset size, high resolution, IR imaging, comprehensive range information, dynamic camera movement, lighting, weather, and background variation, all in a single comprehensive dataset. LRDDv3 instead seeks to capture drones comprehensively by taking range coverage within diverse environments and covers varying altitudes with dynamic lighting conditions. The dataset also leverages 4K imaging, and paired IR data at large scale in a multi-pronged approach that represents a significant advancement over the existing datasets. As such, LRDDv3 positions itself as a comprehensive resource for developing next-generation drone detection systems that can work across the full spectrum of detection ranges and environments.

%% file: sections/3.Dataset.tex
\section{High-Resolution Long-Range Drone Detection Dataset}

\subsection{Dataset Overview}

The primary dataset contribution of our work consists of 101,227 RGB images, all with range information of the distance of the target drone from the camera. Beyond this, we include an additional 1,305 images lacking drone range info to expand weather condition diversity, and add 29,630 thermal images (paired with their RGB counterparts) to enable IR detection. Figure \ref{fig:backgrounds} shows a sample of images from the dataset, featuring a wide variety of backgrounds, camera angles, drone distances, and lighting conditions.

In total, the entire dataset consists of 102,532 RGB images, and 29,630 thermal images, from 17 different data collection days spread over 8 months. The images were sampled at 5 FPS from 128 distinct video clips covering 5.7 hours of video. This sampling rate is distinct, as many other datasets sample at higher rates, or do not subsample video data at all and use the original video frame rate, thus resulting in artificially inflated dataset sizes with limited background diversity. By sampling at 5 FPS, we intentionally reduce the size of the final dataset in favor of image diversity.

In addition to the variation in the overall dataset, we specifically focus on preserving this feature in the test set, using 20k images from 34 video clips taken from a wide array of different locations and weather conditions. Our goal is that this can serve as an effective and comprehensive benchmark for long-range drone detection. Figure \ref{fig:distributions} (a) details the number of images in the official dataset benchmark splits, for both RGB and thermal images.

Figure \ref{fig:distributions} (b) details the distribution of weather (top) and lighting conditions (bottom) in the dataset. A total of 76,965 images were collected in clear weather, 24,356 in rainy weather, and 1,211 in snowy weather. 33,334 images were collected in sunny lighting, 28,183 in partly cloudy lighting, and 41,015 in cloudy lighting.



Figure \ref{fig:distributions} (c) shows the distribution of drone range from the camera. The majority of images include drones at distances from 0-50 meters, but the longest range images include drones at distances of up to 200 meters.

In addition to labeling drones, all instances of birds and airplanes that were present in the images were also annotated, with the final dataset containing a total of 93,652 drone annotations, 6,031 bird annotations, and 722 airplane annotations.

\begin{figure}[!tb] 
\centering
\includegraphics[width=0.99\columnwidth]{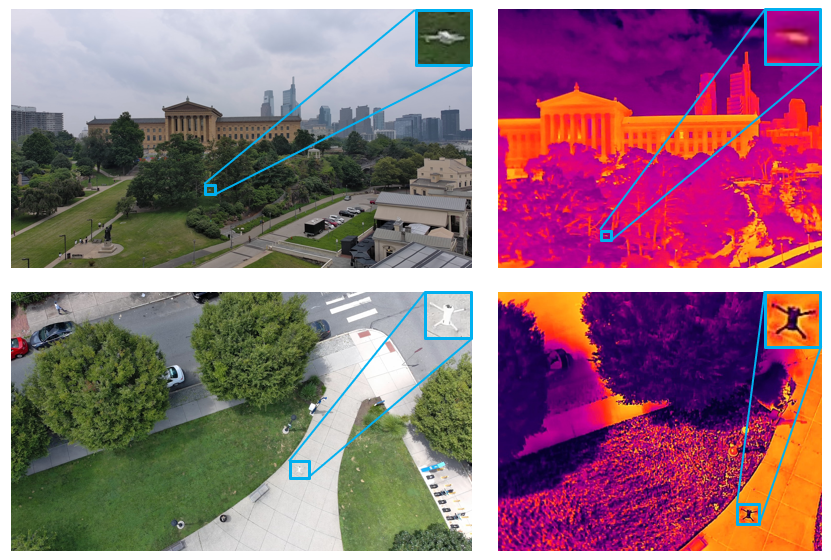}
\caption{Examples of thermal image pairs from the dataset. The RGB images are collected at 4K resolution, and the IR images are recorded at 640x512, centered relative to their RGB counterparts}
\label{fig:thermal_examples}
\end{figure}

\subsection{Data Collection}
Data collection was conducted through controlled drone flights in a range of environments, including urban areas, rural landscapes, forested zones, and open fields. Flights were carried out during various times of the day and under differing weather conditions—sunny, cloudy, and overcast—to ensure diversity in lighting and atmospheric effects.

The dataset is organized into two primary subsets:

\begin{enumerate}
    \item Non-Thermal Subset: Contains RGB images of a target drone captured against a wide range of backgrounds.
    \item Thermal-Pair Subset: Each entry consists of synchronized RGB and thermal images depicting the same scene.
\end{enumerate}

Three drones were used throughout the data collection process: two DJI Mini 3 drones (hereafter, DJI) and one Autel Robotics EVO II Dual 640T V3 (hereafter, Autel-v3). For the non-thermal subset, one DJI drone served as the target drone, while the second DJI acted as the camera drone, capturing 4K video at 30 FPS. In the thermal-pair subset, the Autel-v3 drone—with its dual RGB and thermal imaging system—was used as the camera drone while a DJI drone served as the target drone. The Autel-v3 drone also captured RGB data in 4K at 30 FPS, and recorded IR data at 640x512 at 30 FPS.

The flights were designed to capture the target drone from a wide range of altitudes, angles, and distances. In some sequences, the target drone appeared above the camera drone; in others, it was below or at various lateral angles. The collection process emphasized background diversity, including clear skies, buildings, vegetation, and water bodies (e.g., rivers). To enhance research relevance, particular focus was placed on capturing long-range imagery, including many instances where the target drone is small or difficult to detect within the frame.

Distance variation was a key objective: some sequences show the target drone in close range and clearly visible, while others capture it at extreme distances. Drone metadata, including GPS and altitude logs, were recorded throughout each flight and were used to compute range information for the target drone. 





\subsection{Data Annotation}\label{data_annotation}

\begin{figure*}[!htb] 
\centering
\includegraphics[width=0.9\textwidth]{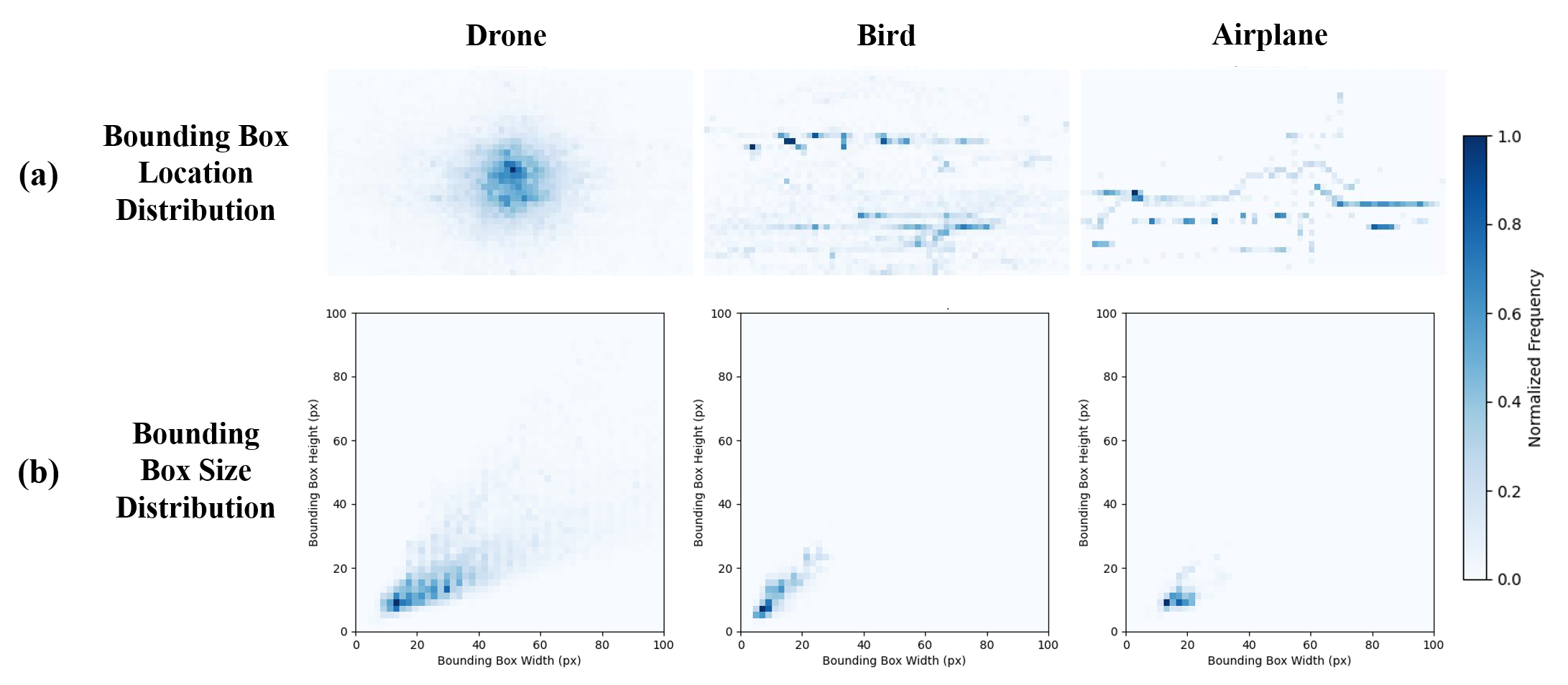}
\caption{(a) Distribution of bounding box locations for the different classes across the dataset. (b) Distribution of bounding box sizes for the different classes in the dataset.}
\label{fig:bbox_distributions}
\end{figure*}

For the final dataset, the raw video data was first manually filtered to remove video clips that were not of sufficient quality for the final dataset. These removed videos were ones where the drone was not in frame for periods of time, or where the drone did not move enough to achieve diverse background information. From the remaining 128 video clips, images were sampled at 5 FPS in order to maintain image and background diversity.

The images were then annotated following a three step process: automated labeling, manual label correction, and label verification. During automated labeling, a YOLOv11 model \cite{yolo11_ultralytics} was used along with the SAHI framework \cite{sahi} to detect drones on the dataset and generate automated prelabels of drone instances. After the prelabels were generated, the data was then manually annotated using the prelabels as a guide to speed up the labeling process. For manual labeling, the open-source software LabelImg \cite{labelimg} was used, as it provides a simple GUI for annotation tasks. Following the manual labeling, the images were finally reviewed to ensure the quality and consistency of the final annotations.

For the thermal images, the annotation process took advantage of the fact that the thermal images and RGB images were collected as pairs, allowing RGB annotations to be converted to apply to the thermal data. As such, once the RGB annotations were completed, a python script was used to automatically rescale and crop the RGB annotations to create the annotation files for the thermal data. These annotations were then also manually reviewed to verify their quality and consistency.

\subsection{Metadata Processing}
\subsubsection{Video-Metadata Matching}
During flight, the drones used for data collection record relevant flight metadata, such as GPS coordinates, altitude, battery power, velocity, control inputs, and video recording state. This metadata was used for determining the range from the camera drone to the target drone in the video frame which we include with all images in the dataset.

In order to match up the camera drone metadata with the recorded video clips, the video recording state variable was used to determine the start of each video and match it with the associated metadata. For matching the target drone metadata to the camera drone metadata, the timestamps in the metadata were used in most cases. However, for some flight days the timestamps were not updated due to lack of internet connectivity, and were incorrect. For these days, the target drone metadata was manually matched up with the camera drone metadata by matching recorded velocity information with observed trajectories in the recorded video. 

This was accurate to within the metadata recording rate of the drone (10Hz), which was faster than the image sub-sampling rate (5Hz) and as such did not significantly contribute to error in the range calculation.


\subsubsection{Range Calculation}

In order to calculate the range information from the camera drone to the target drone, first the horizontal distance between the two drones was calculated from the drone flight GPS coordinates using Haversine formula, where \( \phi_1, \phi_2\) are the latitudes in radians, \( \lambda_1, \lambda_2\)  are the longitudes in radians, and \(R\) is the Earth's radius:

\begin{equation}
a \;=\; sin^{2}\left(\frac{\varphi_{1}-\varphi_{2}}{2}\right)
\end{equation}

\begin{equation}
b \;=\; \cos \phi_1 \cdot \cos \phi_2 \cdot \sin^2\left(\frac{\lambda_{1}-\lambda_{2}}{2}\right)
\end{equation}

\begin{equation}
d_{\text{horizontal}} \;=\; 2R\cdot arcsin\left(\sqrt{ a + b
}\right)  
\end{equation}

The vertical distance between the drones was then calculated from the drone altitude metadata as follows:

\begin{equation}
d_{\text{vertical}} \;=\; |\text{height}_{\text{camera}}-\text{height}_{\text{target}}|
\end{equation}

Then the final distance between the two drones was calculated using both the horizontal and vertical distances.

\begin{equation}
d_\text{range} \;=\; \sqrt{d_{\text{horizontal}}^{2}+d_{\text{vertical}}^{2}} 
\end{equation}

Finally, the drone range was recorded in \texttt{.csv} files for each folder of images, along with the name of the corresponding image file and a \texttt{True}/\texttt{False} flag for if the drone was present in the image.

%% file: sections/4.Benchmark.tex
\section{Dataset Benchmarking}

\begin{table}[!htb]
\caption{YOLOv11 Performance Benchmarks on Det-Fly across various training datasets}
\centering
\label{table:benchmarks}
\begin{tabular}{|l|lll|}
\hline
\textbf{Test Dataset:}    & \multicolumn{3}{c|}{\textbf{Det-Fly}}                                                           \\ \hline
\textbf{Training Dataset} & \multicolumn{1}{l|}{\textbf{mAP@50}} & \multicolumn{1}{l|}{\textbf{mAP@50-95}} & \textbf{F1}    \\ \hline
Drone-vs-Bird             & \multicolumn{1}{l|}{\underline{0.484}}     & \multicolumn{1}{l|}{\underline{0.214}}        & \textbf{0.492} \\ \hline
DUT Anti-UAV              & \multicolumn{1}{l|}{0.431}           & \multicolumn{1}{l|}{0.213}              & \underline{0.485}    \\ \hline
LRDDv1+v2                 & \multicolumn{1}{l|}{0.330}           & \multicolumn{1}{l|}{0.121}              & 0.356          \\ \hline
LRDDv3                      & \multicolumn{1}{l|}{\textbf{0.485}}  & \multicolumn{1}{l|}{\textbf{0.261}}     & 0.468          \\ \hline
\end{tabular}
\end{table}

\subsection{Benchmarks Against Existing Datasets}

To evaluate the effectiveness of our dataset in comparison with other relevant drone detection datasets, we conducted benchmark tests using four different YOLOv11m \cite{yolo11_ultralytics} models trained on our dataset and on three other relevant datasets.

For our training datasets we selected the Drone-vs-Bird Grand Challenge dataset \cite{coluccia_2024_drone_vs_bird} for its scale and similar problem of long-range detection, the DUT Anti-UAV dataset \cite{zhao_2022_antiuav} for its diverse lighting and background conditions, and the previous versions of the Long-Range Drone Detection dataset \cite{rouhi2024_LRDD, rouhi2024_LRDDv2} for their focus on lighting, occlusion, and other real-world challenges. We then evaluate all models on the full Det-Fly dataset \cite{zheng_2021_detfly}, which we chose as a benchmark for its diverse lighting, camera angles, backgrounds, and drone range variation.

In all cases, the models were trained for 50 epochs on an AMD Ryzen Threadripper 3960X 24-Core Processor and two NVIDIA RTX 3090s using a batch size of 32 and an input size of 640x640. All models converged before the end of the 50 epochs, and the best weights were used for inference.

Table \ref{table:benchmarks} shows the results of each model on the three different test datasets, where we report mAP@50, mAP@50-95, and F1 score for each model. As shown in the table, the model trained on Drone-vs-Bird performs the best on F1 score, while the model trained on our LRDDv3 dataset performs best in mAP@50 and mAP@50-95. This verifies that the LRDDv3 dataset is diverse enough to generalize well to other drone detection situations, and performs well in comparison to other related works. Additionally, this result was achieved without leveraging additional data features, such as IR imaging, or the ability to train using higher resolution data.

\subsection{Multi-Resolution Benchmarks}

\begin{table}[!tb]
\caption{Performance of YOLOv11 model variants on the LRDDv3 dataset}
\centering
\label{table:hrd3_benchmarks}
\begin{tabular}{|l|l|l|l|}
\hline
\textbf{Model Variant}    & \textbf{mAP@50} & \textbf{mAP@50-95} & \textbf{F1} \\ \hline
YOLOv11m (640x640)        & 0.543           & 0.288              & 0.482       \\ \hline
YOLOv11m (1920x1920)      & \textbf{0.822}           & \textbf{0.504}              & \textbf{0.800}       \\ \hline
\end{tabular}
\end{table}

Currently, the effective use of information offered by high resolution images is an existing challenge in object detection. Approaches like HRFormer \cite{HRFormer} have sought to resolve this by operating on full image resolution natively, and further research in this area has resulted in the Ultralytics YOLO framework now supporting variable image input size to more effectively use high-resolution images.

As such, we explore the impact of higher resolution training in Table \ref{table:hrd3_benchmarks}, by training a YOLOv11m model at both 640x640 and 1920x1920 input resolutions. In both cases the models were trained on LRDDv3 and tested on the LRDDv3 test set, with the same parameters as the models trained in the previous section, except the 1920 resolution model had the batch size reduced to 8 to accommodate available GPU memory.

As shown in the table, increasing the training input resolution greatly increased all metrics, showing that higher resolution data requires corresponding models that scale with data in order to be leveraged properly. As this is an ongoing area of research, we hope that LRDDv3 can act as an effective testing ground for long-range drone detection that more effectively scales with available data resolution.

%% file: sections/5.Conclusion.tex
\section{Conclusion}

Our work, LRDDv3: High-Resolution Long-Range Drone Detection Dataset, is uniquely positioned for advancing the task of long-distance drone detection through its large dataset size, high resolution, IR imaging, comprehensive range information, dynamic camera movement, lighting, weather, and background variation, all in a single comprehensive dataset. LRDDv3's large size of 102k images and low video sampling rate of 5 FPS ensure varied background diversity in the final dataset. The focus on high resolution 4K imagery and thermal imaging expands current datasets to more readily match the capabilities of modern drone hardware, while comprehensive range information addresses the need for more robust benchmarks in long-range drone detection, allowing for detailed analysis based on range information. Finally, the dynamic camera movement and wide variety of lighting and weather conditions ensure that models remain robust to a wide variety of real-world conditions. While many other datasets exhibit several of these features individually, none capture all aspects, and as such LRDDv3 stands out as a comprehensive resource for the development of drone detection systems that can generalize and remain robust in a wide range of environments and applications.